\documentclass[fleqn,10pt]{SelfArx} 

\usepackage{lipsum} 

\definecolor{color1}{RGB}{0,0,90} 
\definecolor{color2}{RGB}{0,20,20} 

\usepackage{hyperref} 
\hypersetup{hidelinks,colorlinks,breaklinks=true,urlcolor=color2,citecolor=color1,linkcolor=color1,bookmarksopen=false,pdftitle={Title},pdfauthor={Author}}

\JournalInfo{Artificial Intelligence CS 221} 
\Archive{Final Project} 

\PaperTitle{Angrier Birds: Bayesian reinforcement learning}

\Authors{Imanol Arrieta Ibarra\textsuperscript{1}, Bernardo Ramos\textsuperscript{1}, Lars Roemheld\textsuperscript{1}} 
\affiliation{\textsuperscript{1}\textit{Department of Management Science and Engineering, Stanford University, California,  United States}} 

\newcommand{\keywordname}{Keywords} 

\Abstract{We train a reinforcement learner to play a simplified version of the game Angry Birds. The learner is provided with a game state in a manner similar to the output that could be produced by computer vision algorithms. We improve on the efficiency of regular $\epsilon$-greedy Q-Learning with linear function approximation through more systematic exploration in Randomized Least Squares Value Iteration (RLSVI), an algorithm that samples its policy from a posterior distribution on optimal policies. With larger state-action spaces, efficient exploration becomes increasingly important, as evidenced by the faster learning in RLSVI.}

\begin{document}

\flushbottom 

\maketitle 

\tableofcontents 

\thispagestyle{empty} 


\section*{Introduction} 

\addcontentsline{toc}{section}{Introduction} 

Angry Birds has been a wildly successful internet franchise centered around an original mobile game in which players shoot birds out of a slingshot to hit targets for points. Angry Birds is largely a deterministic game: %
\footnote{In our simulation, the physics engine actually seemed to produce semi-random results: the same sequence of moves was not guaranteed to yield the exact same outcomes.} %
a complex physics engine governs flight, collision and impact to appear almost realistically. As such, optimal gameplay could be achieved by optimizing a highly complex function that describes trajectory planning -- instead, we train several reinforcement learning algorithms to play the game, and compare their final performance and learning speeds with that of a human-play oracle.

To simulate gameplay, we adapted an open-source project to recreate Angry Birds in Python \cite{estevaofon} using the Pygame framework. After adapting the base code for our needs, we designed an API which exposed the game as a Markov Decision Process (MDP), largely following the conventions of our class framework.

The Python port we used is a simplified version of the original Angry Birds game: it includes 11 levels, only one (standard) type of bird, only one type of target, and only one type of beam. A level ends whenever there are either no targets or no more birds on the screen. If all targets were destroyed then the agent advances to the next level, otherwise, the agent loses and goes back to the first level. Each level’s score is calculated according to three parameters: number of birds left, shattered beams and columns and destroyed targets. If the agent loses, it incurs an arbitrary penalty. 

This somewhat reduced state-action space allowed a relatively straightforward proof-of-concept -- we believe that very similar algorithms and models would also work for more complex versions of the game.

\section{Model}

\begin{figure*}[ht]\centering 
\includegraphics[width=\linewidth]{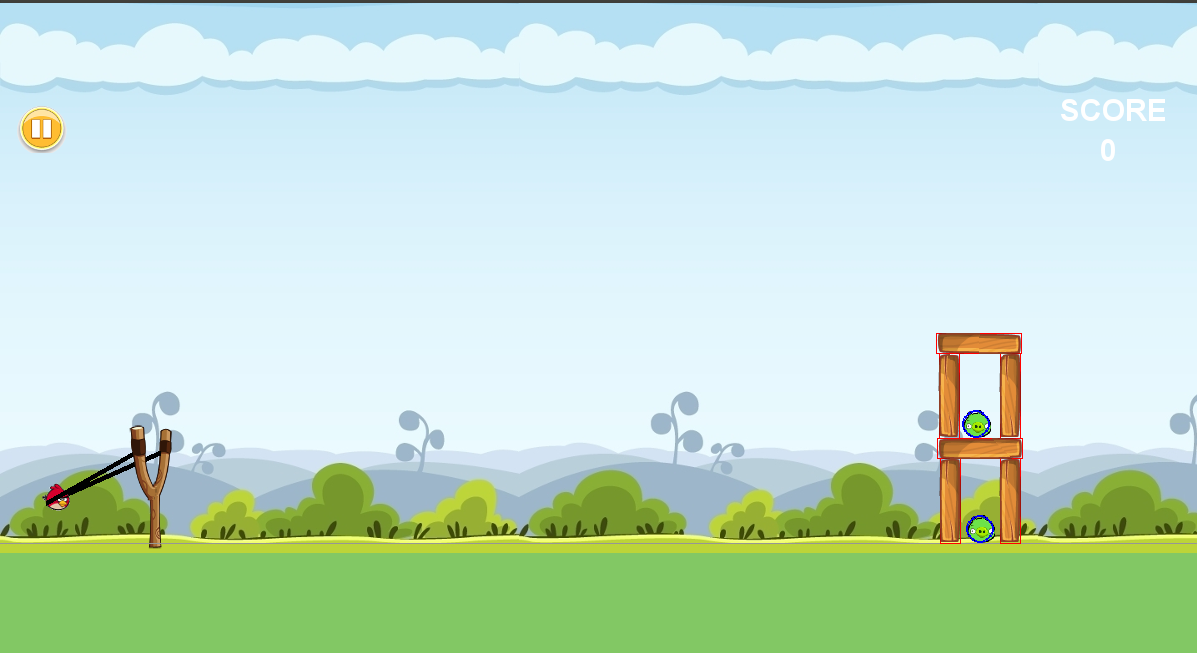}
\caption{Angry Birds gameplay: the stationary slingshot on the left is used to hit the targets on the right. The beam structures can be destroyed, allowing for more complex consequences and interactions of chosen actions.}
\label{fig:view}
\end{figure*}

The game state exposed by our API is composed of the following information. This state representation fully describes the relevant game situation from the player’s perspective and therefore suffices to interface between the learner and the game. We note that all parts of the game state could be obtained from computer vision algorithms, so that in principle no direct interaction with the game mechanics would be necessary.
\begin{itemize}
\item Number of birds left (“ammunition”)
\item Number of targets (“pigs”), and their absolute positions on the game map
\item Number of obstacles (“beams”), and their absolute positions on the game map
\item The game score and level achieved by the player so far.
\end{itemize}

We further simplified game mechanics by making the game fully turn-based: the learner always waits until all objects are motionless before taking the next action. This assumption allowed us to work on absolute positions only.

The set of possible actions is a pair (angle, distance), describing the extension and aiming of the slingshot and thereby the bird's launching momentum and direction. To somewhat accelerate learning, we allowed only shooting forwards. Within that constraint, we allowed all possible extensions and angles -- we compare discretizations of the action space in varying levels of granularity. Our base discretization used 32 different (evenly spaced) actions.


\section{Algorithms}

Reinforcement learning is a sub-field of artificial intelligence that describes ways for a learning algorithm to act in unknown and unspecified decision processes, guided merely by rewards and punishments for desirable and unwanted outcomes, respectively. As such, reinforcement learning closely mirrors human intuition about learning through conditioning, and allows algorithms to master systems that would otherwise be hard or impossible to specify.

A standard algorithm in reinforcement learning is $\epsilon$-greedy Q-learning. While very effective, this algorithm relies on an erratic exploration of possible actions. We implemented a variation which updates a belief distribution on optimal policies, and then samples policies from this distribution \cite{osband2014generalization}. This implies more systematic exploration, since policies that seem unattractive with high confidence will not be re-tried. 

After introducing the two algorithms, we discuss various iterations to find optimal feature extractors.

\subsection{Q-Learning (linear function approximation)}
As exposed in \cite{sutton1998reinforcement}, Q-Learning is a model-free algorithm that approximates a state-action value function $Q_{opt}(s,a)$ in order to derive an optimal policy for an unknown decision process. In order to learn optimal behavior in an unknown game, a Q-Learner needs to both explore the system by choosing previously unseen actions, and to exploit the knowledge thus gathered by choosing the action that promises best outcomes, given previous experience.

Balancing efficient exploration and exploitation is a main challenge in reinforcement learning: the most popular method in literature is the use of $\epsilon$-greedy methods: with probability $\epsilon$, the algorithm would ignore its previous experience and pick an action at random. This method tends to work well in practice whenever the actions needed to increase a payoff are not too complicated. However, when complex sets of actions need to be taken in order to get a higher reward, $\epsilon$-greedy tends to take exponential time to find these rewards (see \cite{osband2014generalization}, \cite{osband2013more}, \cite{osband2015bootstrapped}.)

To aid generalization over the Angry Birds state space, we used linear function approximation: we defined $Q_\mathbf{w}(s,a)=\mathbf{w}^T \mathbf{\phi}(s,a)$, where $\mathbf{w}\in \mathbb{R}^f$, and $\mathbf{\phi}:S\times A\rightarrow \mathbb{R}^f$ is a feature extractor that calculates $f$ features from the given state-action pair. This allows the exploitation step to be an update only on the weights vector $\mathbf{w}$, which we performed in a fashion similar to stochastic gradient descent (following the standard algorithm from \cite{sutton1998reinforcement}).

\subsection{Randomized Least Squares Value Iteration}

Osband et al. propose Randomized Least Squares Value Iteration (RLSVI) \cite{osband2014generalization}, an algorithm that differs from our implementation of Q-Learning in two points:
\begin{enumerate}
\item{Instead of using gradient descent to learn from experience in an online fashion, RLSVI stores a full history of (state, action, reward, new state) tuples to derive an exact "optimal" policy, given the data.}
\item{Given hyperparameters about the quality of the linear approximation, Bayesian least squares is employed to estimate a distribution around the "optimal" policy derived in 1. The learner then samples a policy from this distribution -- this replaces $\epsilon$-greedy exploration.}
\end{enumerate}

Specifically, RLSVI models the state-action value function as follows:
\begin{equation*}
\begin{split}
Q_\mathbf{w}(s_t,a_t) &=\text{Reward}(s_t,a_t) + \gamma \max_{a_{t+1}} Q_\mathbf{w}(\text{Successor}(s_t,a_t),a_{t+1}) \\
& \approx 
\mathbf{w}_t^T \mathbf{\phi}(s_t,a_t) + \nu
\end{split}
\end{equation*}
where $\gamma$ is an arbitrary discount factor, $\nu \sim N(0,\sigma^2)$, and $\sigma$ is a hyperparameter. 

Given a memory of (state, action, reward, new state) tuples, we can use Bayesian least squares to derive $\bar{\mathbf{w}}_t$, the expected value of the optimal policy and $\Sigma_t=\mathrm{Cov}(\mathbf{w}_t)$, the covariance of the optimal policy (the details of which are fairly straight-forward, and can be found in \cite{osband2014generalization}). Note that given the assumption of linear approximation, a weights vector $\mathbf{w}_t$ fully specifies a Q-function, and thereby a policy.

The learner then finally picks a policy by sampling 
\[\hat{\mathbf{w}}_t \sim N(\bar{\mathbf{w}}_t, \Sigma_t)\]
and taking the action such that $Q_{\hat{\mathbf{w}}_t}$ predicts the highest reward. Instead of taking entirely random actions with probability $\epsilon$, the algorithm thus \emph{always} samples random policies, but converges to optimal means as more data is accumulated and variances shrink. Therefore, less time may be expected to be "wasted" on policies that are highly unlikely to be successful.

To keep this algorithm from erratic policy changes when only few observations have been made yet, we initialize with an uninformed prior. Furthermore, inspection showed that $\Sigma$ was almost diagonal. To save computation, we therefore decided to sample using only variances of the individual weights (ignoring covariance between weights).

Osband et al. propose RLSVI in the context of episodic learning, i.e. they suggest updating the policy only at the end of episodes of a fixed number of actions. This helps steady the algorithm -- unfortunately, our game simulation was computationally expensive, and the relatively small number of actions to be simulated forced us to learn in a continuous context. 

\subsection{Feature extractors}	

In designing our feature extractors $\mathbf{\phi}(s,a)$, we followed two premises: first, given that our value function approximation was linear, it was clear that interaction between features would have to be linear; in particular, separating the $x$ and $y$ components of locations would not work. Second, we wanted to give away as little domain knowledge about the game as possible -- the task for the algorithm was to understand how to play Angry Birds without previously having a concept of targets or obstacles.

These two premises effectively impose a constraint on strategies, as there is no reliable way for the algorithm to detect -for example- whether an obstacle is just in front or just behind a target. Regardless, our learner developed impressive performance, as will be seen later. One reason for this may be that the levels in our simulator were relatively simple.

We iterated over the following five ideas to form meaningful features. Due to the relatively complex state-action space, our feature space was fairly big, quickly spanning several thousand features. This turned out to be a problem for RLSVI, as memory and computational requirements grew to be intractable. Rewriting the algorithm to run on sparse matrices offered some help, but ultimately we found reasonable performance with only the best-functioning features.

\subsubsection{Pig Position Values}
In a first attempt, we used rounded target positions as feature values: for every pig $i$ in the given state $S$, we would have feature values $p_i=(x, y, xy)^T$. The features were repeated for every action $a$, and would be non-zero only for the action that was taken. This allows different weights for different position-action combinations, which ultimately implies learning the best position-action combinations.

By including the interaction term $xy$, we had hoped to capture the relative distance of the target to the slingshot (which $x$ and $y$ couldn't establish by themselves in a linear function): this would have allowed fast generalization of target positions. Unfortunately, if in hindsight not surprisingly, this approach failed very quickly and produced practically no learning.

\subsubsection{Pig Position Indicator (PP)}

The next approach was an indicator variable for a fine grid of pig positions: we created a separate feature for every possible pig position and again included the action taken. This created an impractically large feature space, yet worked relatively well. It did, however, clearly "over-fit" -- once a successful series of actions was found, the algorithm would reliably complete a level over and over again. If a target had moved only a few pixels, however, the algorithm would have to be retrained completely.

\subsubsection{Nested Pig Position Counter (NPP)}

\begin{figure}[ht]\centering
\includegraphics[width=\linewidth]{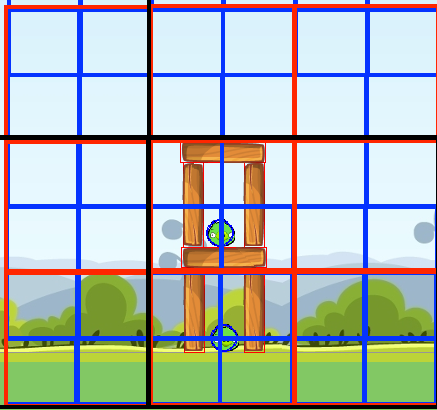}
\caption{NPP and NPPO grid structure: progressively smaller grids create squares within squares, extracting both specific location information and allowing for generalization.}
\label{fig:grid}
\end{figure}

In order to solve the over-fitting problem we developed a nested mechanism that would generalize to unobserved states. To achieve this, we defined 3 nested grids over the game screen as exemplified in Figure \ref{fig:grid}. The three grids are progressively smaller, and each square of each grid is a count of the number of targets contained in it. 

This solved the generalization issue while maintaining some of the nice properties of the previous feature extractor. While the larger grids helped to generalize, the finer ones provided a way to remember when a level had already been observed. 

\subsubsection{Nested Pig Positions: Shifted Counters (NPPS)}

One issue with NPP is that especially for the larger squares, targets that are close to a square boundary are not captured adequately. To improve on this, we created a copy of each grid, shifted diagonally by half a square size. Now every target lay in exactly two squares, allowing to judge whether a target is further to the left or the right within a square.

NPPS was therefore a feature set of two sets of three overlapping square grids each. To our surprise, NPPS performed worse than the simpler NPP. We assume that this is due to the much bigger feature space, and the fact that we could not learn the algorithms until full convergence due to computational limits.

\subsubsection{Nested Pig Positions Counters with Obstacles (NPPO)}
 
Finally, we tried to address the issue of obstacles: as described, we did not want to give away gameplay-related information by representing a fixed relationship between targets and obstacles. We therefore added counters for obstacles in the same way that we used them for targets, hoping that the algorithm would learn to prefer areas with a high $\frac{\text{targets}}{\text{obstacles}}$ ratio. 

Just as in the case of NPPS, we were surprised that adding information about obstacles was detrimental to learning success -- indeed, adding obstacle information stopped learning altogether. As in NPPS, we suspect that NPPO may work well if given more time to converge in the bloated feature space.


\section{Results and Discussion}
The somewhat crude feature extractor NPP provided best results, and RSLVI did indeed outperform the regular $\epsilon$-greedy Q-Learner. RLSVI was particularly impressive for its ability to clear levels at almost the same speed as an (untrained) human player. 

We could not simulate the gameplay until convergence as a result of computational limitations: the game engine we used was not intended for big simulations, and did not even speed up significantly when graphics were disabled.

\subsection{Comparison of feature extractors}
\begin{figure}[ht]\centering
\includegraphics[width=\linewidth]{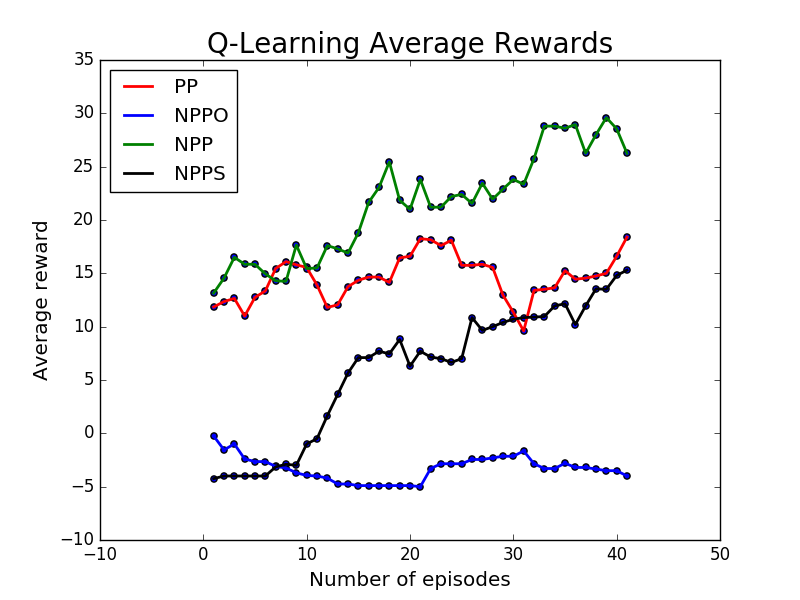}
\caption{Moving Average for the different feature extractors. The dysfunctional "position values" feature extractor has been excluded. We display a moving average game score over the \emph{next} 10 attempts after a number of attempts has been made (where an attempt is defined as any number of shots until a level is failed). We successively used one attempt to explore, and then recorded the following attempt without exploration, in order to not distort results with exploration behavior. The y-axis is in units of $10^4$ points.}
\label{fig:QAverageReward}
\end{figure}

We compared our five different feature extractors on the Q-Learning baseline algorithm, as displayed in figure \ref{fig:QAverageReward}.

It is noteworthy that PP starts on a high baseline score, but does not improve much from there: this is due to the fact that the learner very quickly masters a level, but then does not generalize to following levels. As a consequence, PP tends to get stuck and achieve relatively constant game scores in the first few levels.

In contrast, NPP learns to master easy levels equally fast, but then carries over to harder levels better, yielding higher total scores per attempt. 

NPPS shows a very similar slope to NPP, supporting the explanation that the sheer number of features slows the learning process down. However, NPPS is clearly outperformed by NPP within our simulation time frame. We are somewhat surprised at the complete failure of NPPO to improve over time.

\subsection{RLSVI vs. regular Q-Learning}

\begin{figure}[ht]\centering
\includegraphics[width=\linewidth]{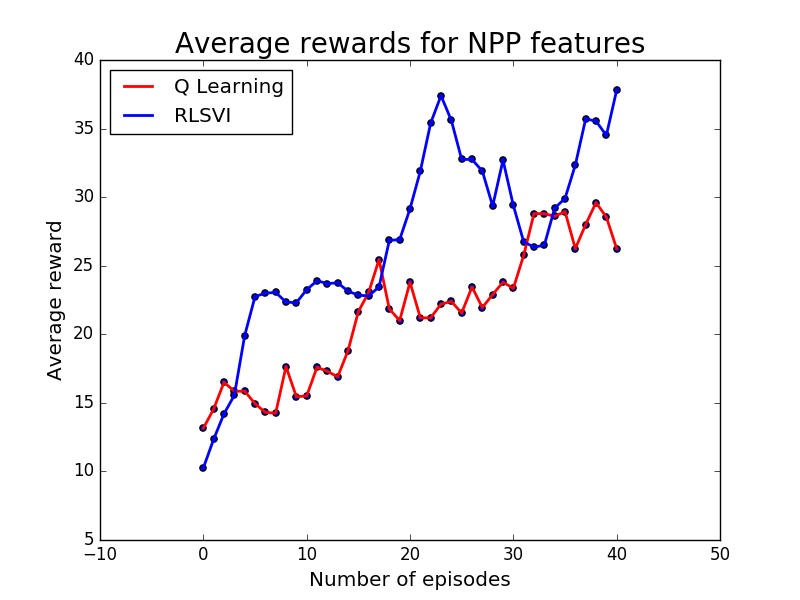}
\caption{RLSVI is faster and achieves greater rewards than Q-Learning. We display a moving average game score over the \emph{next} 10 attempts after a number of attempts has been made (where an attempt is defined as any number of shots until a level is failed). We successively used one attempt to explore, and then recorded the following attempt without exploration, in order to not distort results with exploration behavior. The y-axis is in units of $10^4$ points.}
\label{fig:NPPAverageReward}
\end{figure}

Exploring the state-action space according to posterior beliefs did indeed produce better results than the $\epsilon$-greedy Q-Learner. We trained both algorithms on the same feature extractor (NPP) and compared a moving average game score per attempt. RLSVI achieved higher scores overall in the simulated time frame, and learned more quickly. A comparison of the two algorithms is given in figure \ref{fig:NPPAverageReward}.

It should be noted that in our implementation, RLSVI required significantly more memory and computation than the regular Q-Learner, since it required keeping a complete history of observed features, and matrix operations on large data blocks.

\subsection{Learning success}

Neither a human player -our oracle for score comparison- nor any of the algorithms managed to go through all 11 levels provided by the game engine we used in one attempt. Both the regular Q-Learner and RLSVI however outperformed the human player (who shall remain unnamed, given that he was shamefully beat by crude algorithms) in terms of maximum points achieved in a single attempt, and in terms of the highest level reached. Table \ref{tab:max_score} summarizes the maximum score attained and the highest level reached for different players.

\begin{table}[hbt]
\centering
\begin{tabular}{llrr}
\toprule
\multicolumn{2}{c}{Model} & \multicolumn{2}{c}{Maximum Performance}\\
\midrule
Algorithm & Features & Score & Level \\
\midrule
Human & ---- & $210000$ & $7$  \\
Q-Learning & PP & $170000$ & $8$ \\
Q-Learning & NPP & $495000$ & $9$ \\
Q-Learning & NPPO & $120000$ & $3$  \\
RLSVI & NPP & $550000$ & $8$ \\
\bottomrule
\end{tabular}
\caption{Maximum level and scores achieved. Both the regular Q-Learner and RLSVI outperform our untrained human player.}
\label{tab:max_score}
\end{table}

Comparing the number of attempts required to pass a given level provides interesting insights: especially in later levels, RLSVI's exploration proves very efficient, as it passes the levels at almost the same number of attempts as required by the human player. With the same features, regular Q-Learning required 3 times as many attempts, as can be seen in table \ref{tab:num_trials}.

\begin{table}[hbt]
\centering
\begin{tabular}{llrrrrr}
\toprule
\multicolumn{2}{c}{Model} & \multicolumn{5}{c}{Number of trials to finish:}\\
\midrule
Algorithm & Features & Level 0 & Level 5 & Level 7 \\
\midrule
Human & ---- & $1$ & $2$ & $4$ \\
Q-Learning & PP & $2$ & $5$ & $10$ \\
Q-Learning & NPP & $1$ & $4$ & $15$ \\
Q-Learning & NPPO & $4$ & $20$ & $40$ \\
RLSVI & NPP & $1$ & $2$ & $5$ \\
\bottomrule
\end{tabular}
\caption{Average number of trials needed to finish a level. RLSVI learns very quickly in later levels, and passes levels after impressively short exploration periods.}
\label{tab:num_trials}
\end{table}

\section{Conclusions and Future Work}

RLSVI's Bayesian approach to state-action space exploration seems like a promising and rather intuitive way of navigating unknown systems. Previous work has shown Bayesian belief updating and sampling for exploration to be highly efficient \cite{osband2013more}, which our findings confirm. RLSVI beat both the baseline Q-Learning algorithm and, somewhat embarrassingly, our human oracle.

A main constraint on our work was simulation speed, limiting the algorithm convergence we could achieve. We would have liked to further explore possible features, especially after more computation. 

In the same vein, it would be interesting to train a deep neural network for the $Q(s,a)$ function, to allow learning more complex interaction effects, in particular between targets and obstacles. Promising results combining Bayesian exploration with deep reinforcement learning have been shown in \cite{osband2015bootstrapped}.

Finally, it may be interesting to explore different distribution assumptions within RLSVI; assuming least-squares noise to be normally distributed, and therefore sampling from a normal distribution around expected optimal policies may be particularly prone to get stuck in local minima. Using bootstrapping methods in lieu of closed-form Bayesian updates may prove to be a powerful improvement on the RLSVI algorithm explored here.

\section*{Appendix} 
Our work can be found at \\ \url{github.com/imanolarrieta/angrybirds}. Original codes for Angry Birds in Python can be found at \url{https://github.com/estevaofon/angry-birds-python}.

\section*{Acknowledgments} 
We would like to thank the CS221 teaching team for an inspiring quarter.

\phantomsection
\bibliographystyle{unsrt}
\bibliography{sample}
\nocite{*}


\end{document}